\documentclass{article}
\usepackage{ijcai24}

\usepackage{times}
\usepackage{soul}
\usepackage{url}
\usepackage[hidelinks]{hyperref}
\usepackage[utf8]{inputenc}
\usepackage[small]{caption}
\usepackage{graphicx}
\usepackage{amsmath}
\usepackage{amsthm}
\usepackage{booktabs}
\usepackage{algorithm}
\usepackage{algorithmic}
\usepackage[switch]{lineno}

\title{Federated Learning Optimization: A Comparative Study of Data and Model Exchange Strategies in Dynamic Networks}

\author{
Alka Luqman$^1$
\and
Yeow Wei Liang Brandon$^1$\And
Anupam Chattopadhyay$^1$
\affiliations
$^1$Nanyang Technological University, Singapore\\
\emails
\{alka001, yeow0070, anupam\}@ntu.edu.sg
}

\begin{document}

\maketitle

\begin{abstract}
The promise and proliferation of large-scale dynamic federated learning gives rise to a prominent open question - is it prudent to share data or model across nodes, if efficiency of transmission and fast knowledge transfer are the prime objectives. This work investigates exactly that. Specifically, we study the choices of exchanging raw data, synthetic data, or (partial) model updates among devices. The implications of these strategies in the context of foundational models are also examined in detail. Accordingly, we obtain key insights about optimal data and model exchange mechanisms considering various environments with different data distributions and dynamic device and network connections. Across various scenarios that we considered, time-limited knowledge transfer efficiency can differ by up to \textbf{9.08\%}, thus highlighting the importance of this work.
\end{abstract}

\section{Introduction}
Federated Learning (FL) is a decentralized learning paradigm that has gained significant attention recently due to its potential to address privacy concerns and data ownership issues in large-scale machine learning applications. However, the dynamic nature of the participating devices and the complexities introduced by large-scale, heterogeneous data distributions pose significant challenges to the effectiveness and efficiency of FL systems. 
The amount of data collected daily is about 2,500,000 Terabytes and it is increasing in volume every day \cite{marr2023}. In 2020, a single self-driving car was collecting up to 5TB of data a day \cite{miller_2020}. We are using more sensors and cameras to collect data today thanks to the Internet of Things (IoT), wearable personal devices, connected autonomous vehicles etc. The increasing infeasibility of processing this amount of data on a single machine, combined with the private nature of the captured data, makes FL an apt technique for creating application-specific ML models. 
Several studies have investigated various aspects of FL, including communication efficiency, privacy preservation, and model convergence \cite{DBLP:journals/corr/abs-1912-04977}. Recent advancements have explored the use of synthetic data as an alternative to raw data exchange \cite{goetz2020federated,hu2022fedsynth}, aiming to mitigate privacy risks and reduce communication overhead. An optimal strategy for exchanging knowledge among devices in large dynamic settings remains an open research question.

\section{Literature Review}
Distributed machine learning (DML) distributes the computational workload of ML algorithms across multiple machines \cite{DBLP:journals/corr/abs-1912-04977}. This intuitively channels model parallelism, where the ML model or different parts of one ML model are computed across various devices. Data parallelism, on the other hand, distributes the dataset across devices such that each device operates on a smaller batch of data. This is suitable for situations with large amounts of data \cite{https://doi.org/10.48550/arxiv.1404.5997}. FL follows such a process of data parallelism to train ML models on data present at various edge devices, which then share their model updates (weights or gradients) to be aggregated by a trusted parameter server. In the absence of a trusted central aggregator, decentralized coordination can be performed using Peer-to-Peer FL (P2P FL).\par 

\cite{DBLP:journals/corr/McMahanMRA16} state that to optimize FL algorithms like FedAvg, the communication costs significantly outweigh computation costs and that computational costs become negligible with a sufficiently large number of clients. In a realistic large-scale peer-to-peer network, especially in IoT, these communications occur over networks where problems like low client availability, fluctuating device power, and dynamic network bandwidths are expected. In such dynamically changing environments, it becomes imperative that the ML infrastructure also adapts dynamically, decoupled from the objective of the ML model. \par

In order to do this, a device participating in training ML models can opt to implement a DML approach or an FL approach at any point in time. It has the flexibility to share raw data or model updates with its trusted peers to create an aggregated global model. Such a model can be extended to untrusted peers via \cite{fung2020mitigating,NEURIPS2021_7d38b1e9,10.1145/3383455.3422562} to mitigate attacks against robustness and privacy. In the case of privacy-sensitive ML applications, devices can opt to share synthetic data generated from the raw data they have access to. This approach has been demonstrated by \cite{goetz2020federated,hu2022fedsynth} to be equally effective as sharing raw data. \par  

\begin{figure*}[ht]
    \centering
    \includegraphics[scale=0.65]{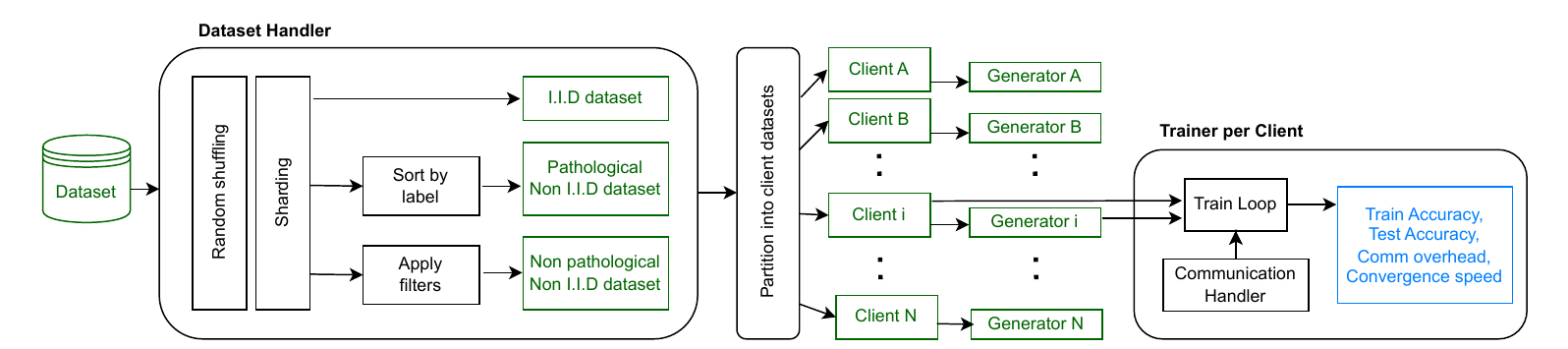}
    \caption{Workflow depicting Dataset Handler which utilizes an array of shuffling, sharding and image transformations to systematically generate realistic client datasets. Each client trains its own generator model to prevent data leakage while the communication handler dynamically adopts between data-sharing strategies and model-sharing strategies.}
    \label{fig:workflow}
\end{figure*}

The FL approach may be extended to pre-trained foundational models in order to combine the strengths of pre-training with the privacy benefits of federated learning and adaptability to client-specific distributions. Deploying multiple instances of such large models and fine-tuning the entire models can incur significant storage and communication costs. A straightforward way to reduce this cost is to perform low-rank decomposition on the pre-trained weight matrix using LoRA\cite{hu2021lora} to split it into frozen layers and trainable layers.
Adapter-tuning\cite{houlsby2019parameter} and parameter-efficient fine-tuning\cite{ding2023parameter,he2022parameter} methods can be adopted to make the model updates more efficient for FL.\par 

Here we develop an algorithm that identifies the optimal action a device in a peer-to-peer network can take to enhance communication efficiency without sacrificing model accuracy. It considers essential factors such as the size of other connected peers' datasets, their accuracy, CPU speed, available RAM, bandwidth, number of connected peers, and privacy constraints related to data sharing. \par

\section{Methodology}
In this work, we first want to evaluate the performance of models in dynamic networks and characterized by varying non-i.i.d data distributions. The results of these comparative studies will then be used to create an adaptive algorithm that implements a hybrid transfer model using data and/or model sharing techniques.\par
\subsection{Dataset \& Preprocessing}
The dataset used for benchmarking was CIFAR-10, which comprises 60,000 32x32 color images in 10 classes, with 6,000 images per class. The training dataset of 50,000 images was partitioned among multiple devices to simulate a large dynamic federated learning environment.\par
\cite{https://doi.org/10.48550/arxiv.1806.00582} identified that the test set accuracy can fall by as much as 55\% in a highly skewed scenario where each client only has one class. As this is a highly probable scenario in FL, the following standard methods from \cite{DBLP:journals/corr/abs-1912-04977} were used to create non-i.i.d data distribution in our experiments.

\begin{algorithm}[H]
    \caption{Pre-Communication Algorithm}
    \label{alg:pre-communication}
    \textbf{Input}: $self$.K\_degree $> 0$\\
    \begin{algorithmic}[1] 
        \FOR{client \textbf{in} network\_graph}
        \STATE peers $\gets $BFS($self$, $self$.K\_degree) 
        \STATE action\_score\_tuples $\gets$ Ranking($self$, peers)
        \STATE action\_queue $\gets$ PriorityQueue(action\_score\_tuples)
        \STATE costs, actions\_to\_execute $\gets$  0, empty list
        \WHILE{costs $< self$.bandwidth}
        \STATE action $\gets$ action\_queue.$pop$()
        \IF{action.cost $>$ ($self$.bandwidth $-$ costs)}
        \STATE break
        \ELSE
        \STATE actions\_to\_execute.append(action)
        \STATE costs $+=$ action.cost
        \ENDIF
        \ENDWHILE
        \ENDFOR
        \STATE \textbf{return} actions\_to\_execute
    \end{algorithmic}
\end{algorithm}

\begin{itemize}
    \item \textbf{Prior probability shift} - It occurs across clients $i$ and $j$ when the set of labels observed by client $i$ may not be the same as the set observed by client $j$. This is a label distribution skew where $P(y_i) \neq P(y_j)$ although $P(x|y)$ is the same across clients. This is referred to as pathological non-i.i.d data set distribution in \cite{DBLP:journals/corr/McMahanMRA16}.
    \item \textbf{Quantity skew} - It occurs when one client has more data than another. It is implemented through a sharding function to split the dataset into preset ratios across a specific number of clients. This function assumes that the data is already shuffled or sorted for i.i.d data and pathological non-i.i.d data.
    \item \textbf{Co-variate shift} - It occurs when the features representing the same label across client $i$ and $j$ vary. This is when $P(x_i) \neq P(x_j)$ although $P(y|x)$ is the same. This can typically occur when one camera is slanted or has a different lens quality. It can also occur when clients belong to different geographical regions, having different dialects of speech or styles of writing etc. \textbf{Concept drift} is a similar but related concept which occurs when the distribution of labels $y$ is the same, yet the distribution of $x$ shifts from client to client across class labels. As this is an alternate methodology to introduce non-i.i.d behaviour, we refer to this as non-pathological non-i.i.d data set distribution. This is implemented by applying common image data augmentation techniques like translation, random crop, image rotation, application of Gaussian noise, distortion etc. 
\end{itemize}

\subsection{Federated Learning Framework}

\cite{DBLP:journals/corr/abs-1901-11173} and \cite{9431011} introduced FL as a graph learning problem using a peer-to-peer architecture. There is no single global model in the network, but nodes aggregate models from their peers and eventually converge to a global optimum. We implemented a federated learning framework using PeerFL\cite{luqman2024peerfl} to create a simulation of peer-to-peer federated learning scenarios with multiple clients joining dynamic adhoc networks. Each client receives a subset of the preprocessed dataset, and the partitions are updated dynamically to simulate real-world scenarios where devices may join or leave the network. \par
The computation cost at device $i$ is defined as a function of its battery level $D_{i}^{pow}$ and available memory $D_{i}^{mem}$ to perform on-device training.
\begin{equation}
\begin{aligned}
\mathcal{C}^{compute}_{i}=f(D_{i}^{pow},D_{i}^{mem})= D_{i}^{pow} * D_{i}^{mem}
\end{aligned}
\label{eq:compute_cost}
\end{equation}
The communication cost is calculated as a function of the channel bandwidth $E_{i}^{bw}$ and the size of the message being transmitted over it $S$. 
\begin{equation}
\begin{aligned}
\mathcal{C}^{comm}_{i}=g(E_{i}^{bw},S)= \frac{\alpha S}{E_{i}^{bw}}
\end{aligned}
\label{eq:comm_cost}
\end{equation}
The simulation parameter ranges are sourced from \cite{ng2021reputation}.\par

Assuming each client state is represented by a set of information and the other connected clients can access that information by traversing the graph, a three-step algorithm is proposed in Algorithm \ref{alg:pre-communication} for each client to discover the best actions to take with its $k$th degree peer given constraints. The constraints are further probed using a ranking function which can be modified to represent any characteristic in the node such as the local data set quality, CPU availability for offloading training, availability of GPU, etc. This step initiates the network exploration and peer prioritization for the model training.\par

\begin{algorithm}[!ht]
\caption{\textbf{\textit{BFS}}: Breadth-first Search to $k$th degree for Client Information Retrieval}\label{alg:BFS}
\textbf{Input}: $client$, history, k$=-1$\\
\begin{algorithmic}[1]
    \IF{k$ = -1$}
        \STATE k$\gets client$.K\_degree  
        \STATE peer\_list $\gets empty list$
        \STATE peer\_queue $\gets [client, $k$]$  
    \ENDIF
    
    \WHILE{peer\_queue is not empty}
        \STATE peer, current\_k $\gets$ peer\_queue.$pop$(0)
        \STATE peer\_list.append(peer)
    
        \IF{current\_k $== 0$}
            \STATE continue
        \ENDIF
        
        \FOR{$child$ \textbf{in} peer.peers}
            \IF{$child$ \textbf{in} peer\_list}
                \STATE continue
            \ENDIF
            \STATE peer\_queue.append(($child$, k-1))
        \ENDFOR
    \ENDWHILE
    \STATE \textbf{return} peer\_list
\end{algorithmic}
\end{algorithm}

\begin{algorithm}[!ht]
\caption{\textbf{\textit{Ranking}}: Client-level Communication Decision Algorithm}\label{alg:Comm_Decision}
\textbf{Input}: $client$, peers\\
\begin{algorithmic}[1]
    \STATE actionsToRun $\gets empty list$
    \FOR{peer in peers}
        \IF{peer.accuracy $> client$.accuracy}  
            \STATE runFedAvg $\gets$ True
            \FOR{recent\_actions in $client$.action\_history}
            \IF{recent\_actions \textbf{contains} FedAvg}
                \STATE runFedAvg $\gets$ False
            \ENDIF
            \ENDFOR
            \IF{runFedAvg}
                \STATE reward $\gets$ Mean($client$.accuracy, peer.accuracy)
                \STATE actionsToRun.add($runShareModel$($self$, peer),  reward)
            \ENDIF
        \ENDIF
        \STATE
        \IF{$client$ in peer.trustedPeers}
            \STATE dataset\_type $\gets$ Synthetic
        \ELSE
            \STATE dataset\_type $\gets$ Sampled
        \ENDIF
        \STATE reward $\gets$ scoreContext($client$, peer)
        \STATE actionsToRun.add($runShareData$($client$, peer, dataset\_type), reward)
    \ENDFOR
    \STATE \textbf{return} actionsToRun
\end{algorithmic}
\end{algorithm}

The mechanism chosen for network exploration is a breadth-first search. At each client we apply Algorithm \ref{alg:BFS} recursively to traverse and retrieve parameters of all connected peers to the $k$th degree. Limiting the peer exploration to a finite $k$ neighbours helps prevent the algorithm from getting stuck in infinite loops in the case of cyclic graphs, and is memory-efficient in the case of very large graphs. Algorithm \ref{alg:Comm_Decision} is the communication handler, which decides the action to be performed based on the characteristics of peers and clients.\par
The complexity of Algorithm \ref{alg:BFS} is $O(K\cdot(V+E))$, influenced by $V$, the number of peers and $E$, the connections between peers, limited by $K$ which is the maximum degree of exploration. The ranking operation in Algorithm \ref{alg:Comm_Decision} contributes complexity of $O(V\log V)$. Therefore the overall complexity of Algorithm \ref{alg:pre-communication} is $O(K\cdot(V+E) + V\log V)$. While this performs well in most cases, the density of device connections has a significant impact in the worst case. Increasing the exploration depth $k$ leads to a higher communication cost in practice in dense graphs (with average node degree $> \frac{2E}{V}$), even though the asymptotic time complexity remains comparable to sparse network graphs.\par

\subsection{Data \& Model Exchange Strategies}
\label{sec:strategies}
\begin{enumerate}
    \item \textbf{Exchanging Raw Data:}
In this approach, the raw data samples are exchanged among the peers.
Each client trains a local model on its dataset and sends the updated model to its peer or cluster head.
This device aggregates the model updates to with its own local model and propagates this further to the rest of the network.
\item \textbf{Exchanging Synthetic Data:}
Synthetic data generation techniques, such as Variational AutoEncoders \cite{https://doi.org/10.48550/arxiv.1312.6114} and  Vector-Quantisation Variational Autoencoders \cite{https://doi.org/10.48550/arxiv.1711.00937}, are used to create artificial data samples based on the local data of each client. VAEs are chosen specifically since they create lightweight models suitable to run on hardware-constrained devices. The synthetic data samples are then exchanged among the peers, and the federated learning process proceeds similarly to the raw data exchange approach. 
\item \textbf{Exchanging Model Updates:}
Instead of exchanging raw or synthetic data, only the model updates (i.e. gradients) are exchanged among the devices.
Each client computes the model gradients based on its local dataset and sends these gradients to be aggregated via FedAvg\cite{mcmahan2017communicationefficient} and processed further.
\item  \textbf{Using Foundational Models:}
This method is a variation of strategy 3, i.e. exchanging model updates, but the models used are foundational models which are further fine-tuned at each client and the model weights of trainable parameters are aggregated via averaging. We experimented with YOLOv8\cite{Jocher_Ultralytics_YOLO_2023} models pretrained on COCO and applied to object detection on the KITTI dataset\cite{Geiger2013IJRR}.
\end{enumerate}

Each client, apart from storing its private data and model, also maintains its own synthetic data generator, trained on its respective dataset partition. This generator is used to craft samples instead of sharing raw data samples when the network is untrusted. This generation is performed on the fly at each client to preserve device-level privacy. Backbone models of VGG16\cite{https://doi.org/10.48550/arxiv.1409.1556} and SqueezeNet\cite{https://doi.org/10.48550/arxiv.1602.07360} are compared using initial learning rates of 0.01 and 0.001 respectively on batch sizes of 256. \par

\subsection{Metrics}
To evaluate the performance of the different data exchange strategies, we considered the following metrics:
\begin{enumerate}
    \item Model Accuracy: The accuracy of the best model at each device on the test dataset.
    \item Communication Overhead: The amount of data exchanged between the clients and the central server.
    \item Convergence Speed: The number of communication rounds required for the model to converge.
\end{enumerate}
For each peer we formulate the resultant ML model accuracy as a reward that each choice of exchange strategy results in.
\begin{equation}
\begin{aligned}
    Reward(Action) = E(AccuracyIncrease|Action)
\end{aligned}
\label{eq:reward}
\end{equation}
This score is weighted by the cost of communication (Eq \ref{eq:comm_cost}), which is the amount of data each action costs.

For object detection on KITTI, MAP is evaluated with IoU threshold of 50\%.
The experiments were designed to compare the performance of the different data exchange strategies under varying conditions, such as different network topologies and communication frequencies. Each experiment was repeated multiple times to ensure the robustness and reliability of the results.

\begin{figure}[!t]
    \centering
    \includegraphics[scale=0.2]{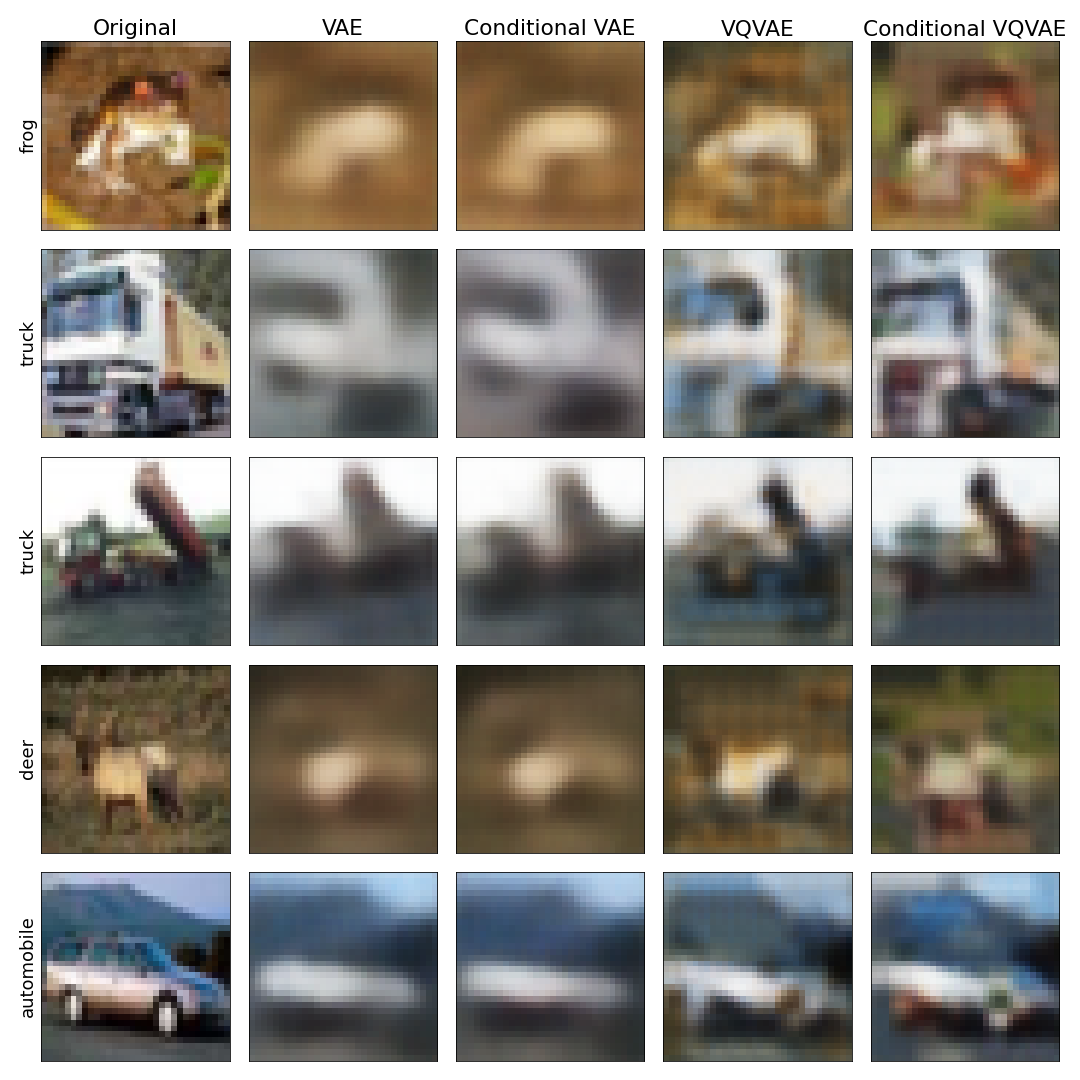}
    \caption{Synthetically generated samples of 5 images across classes of VAEs used in the Generator.}
    \label{fig:VAE Sampling of 5 Images}
\end{figure}

\section{Experimental Studies}
\begin{table}[b]
    \centering
    \begin{tabular}{cc}
    \toprule
        \textbf{Model backbone} & \textbf{Model packet size (MB)}\\
    \midrule
        SqueezeNet & 7.396\\
        VGG16 & 60.771\\
        YOLOv8 & 27.890\\
   \toprule
        \textbf{Client's percentage} & \textbf{Data packet} \\
        \textbf{of dataset} & \textbf{size (MB)}\\
    \midrule
        10\% & 58.785\\
        20\% & 117.569 \\
        30\% & 176.354 \\   
        60\% & 352.707 \\
        80\% & 470.276\\
    \bottomrule
    \end{tabular}
    \caption{Average size of compressed model weight packets and data packets (in MB) during transmission based on which Communication Handler chooses a strategy.}
    \label{tab:weights_size_data_size}
\end{table}

\begin{table*}[!h]
    \centering
    \begin{tabular}{ccccccc}
    \toprule
        \textbf{Data Distribution} & \textbf{Strategy 1} & \textbf{Strategy 2} & \textbf{Strategy 3} & \textbf{Strategy 4} & \textbf{Most used}& \textbf{Ideal} \\
        \textbf{at clients} & \textbf{only (Acc\%)} & \textbf{only (Acc\%)} & \textbf{only (Acc\%)} & \textbf{only (MAP\%)} & \textbf{strategy} &\textbf{for}\\
    \midrule
        I.I.D & 95.3 & 95.1 & 93.1 & 76.7 &  & \\
        \midrule
        Pathological & 85.6 & 85.1 & 70.1 & 51.7 & \textbf{1,2} & High idle \\
        non I.I.D & &  &  &  & & time  \\
        \midrule
        Non Pathological & 82.8 & 89.2 & 83.0 & 71.6 & \textbf{2,4}& High local \\
        non I.I.D &  &  &  &  & & compute \\
        \midrule
        Quantity skew & 94.6 & 97.2 & 72.8 & 64.5 & \textbf{1,2 first}\\
        \midrule
        $<$5\% data with quantity & 54.3 & 53.1 & 66.4 & 38.2 & \textbf{3 first} & Dense\\
        skew at each client &  &  &  &  &  & network graph\\
    \bottomrule
    \end{tabular}
    \caption{Consolidated insights on optimum choice of strategy from Section \ref{sec:strategies}.}
    \label{tab:results_insights}
\end{table*}

\begin{figure*}[!h]
    \centering
    \includegraphics[width=0.55\textwidth, angle = -90]{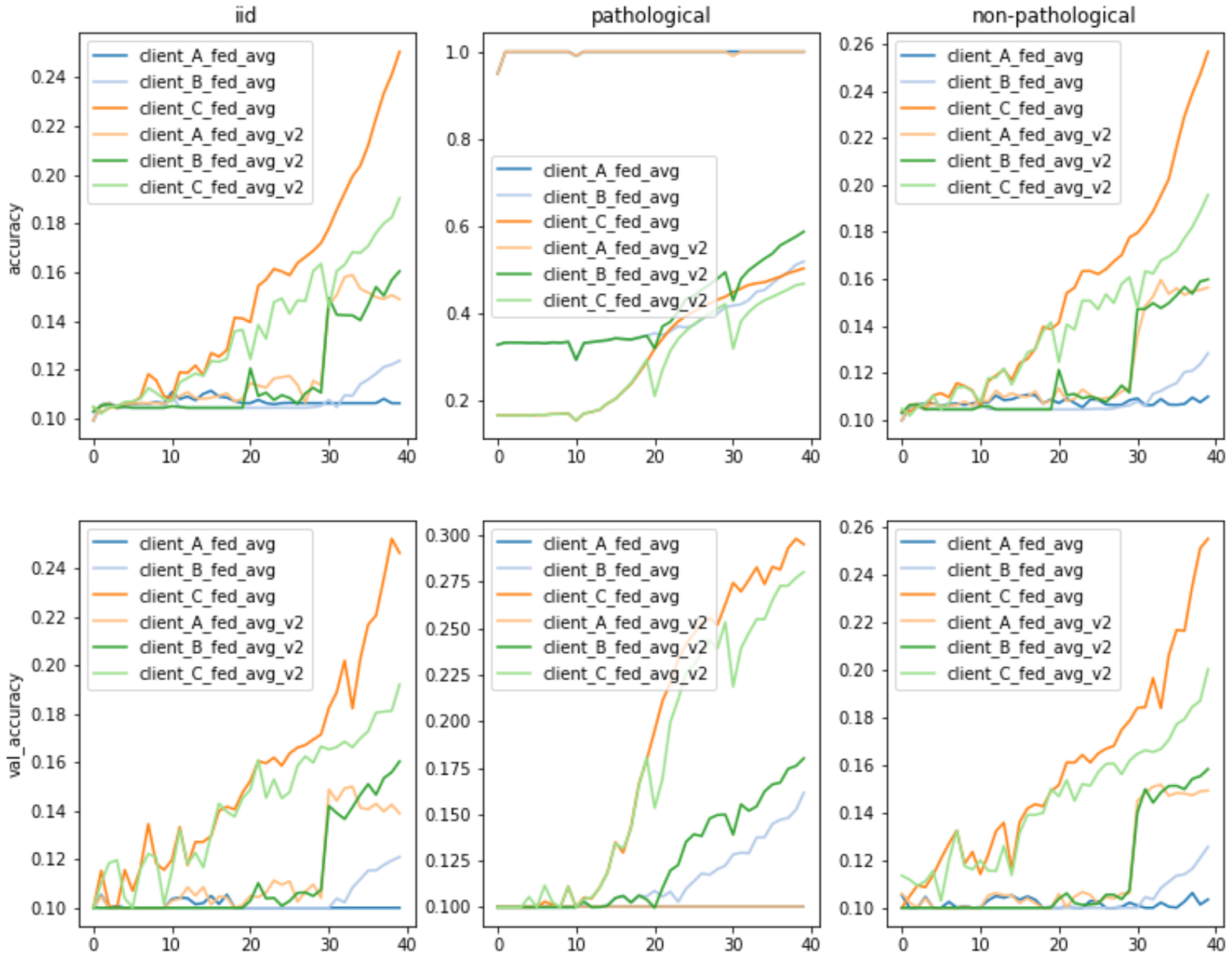}
    \caption{Results of 3 clients trained from scratch using model sharing strategy. Pathological non-i.i.d distribution seems to worsen all FedAvg, mostly due to missing classes being difficult to learn across multiple clients simultaneously. For client C however, it performs worse consistently across all distributions. Model sharing seems to contribute most when data samples are low and the data set may be more difficult, as seen from the larger improvement.}
    \label{fig:3 Client FedAvg Comms Count Comparison From Scratch}
\end{figure*}

The experiments were conducted to evaluate the performance of different data exchange strategies in large-scale dynamic federated learning settings. The experimental setup is inspired by the methodologies used in previous studies on federated learning and data exchange strategies \cite{mcmahan2017communicationefficient,bonawitz2019towards} and presented in Figure \ref{fig:workflow}. \par

Both VGG16 and SqueezeNet performed considerably better when initialized using pre-trained weights from tiny ImageNet, as opposed to random initialization and training only on the local datasets. For all experiments excluding SqueezeNet, the initial probing rounds were leveraged to also perform model sharing, to make maximum use of the bandwidth consumed. This practice resulted in locally converged models, the accuracy of which could be later improved using other exchange strategies.\par

A stark contrast in quality across generator models was observed by utilizing a discrete latent space (VQVAE, CVQVAE) over a Gaussian latent space (VAE, CVAE). A few synthetically generated samples are visualized in Figure \ref{fig:VAE Sampling of 5 Images}. Conditional VAEs are cheaper to run compared to VQVAEs which require a larger model to generate noise for sampling, and thus are suited for memory-constrained devices.  \par

Table \ref{tab:weights_size_data_size} shows that sharing data consumes a higher bandwidth cost but takes fewer communication rounds to converge while model weights incur a lower bandwidth cost but require multiple successive communication rounds to improve. The general trend noticed is that model-sharing strategies create a \textit{stable} model while data-sharing strategies have \textit{faster} convergence. Performing model-sharing first shows the ability to pull a model out of local minima with continuous adoption of data-sharing strategies in later stages.\par
With pre-trained foundational models, the law of diminishing effects is observed and sharing raw or synthetic data mostly leads to overfitting, and the clients benefit better from model sharing. But on clients that have pathological non-I.I.D data, model sharing is rarely seen to be a good choice. They mostly benefit from data sharing, a sample subset of the raw data from trusted peers, or synthetic data in the case of non-trusted peers. The insights from our experiments are tabulated in Table \ref{tab:results_insights} for easy inference. \par

A surprising observation was that sharing synthetic data was as effective as using pre-trained models to improve accuracy. This is attributed to two reasons - the imbalanced data distribution among the devices, introduced through quantity skew, and the distributional bias introduced in the raw data by the co-variate shift operations.\par
Learning from indirect data representations in highly non-I.I.D settings provides a semblance of privacy since the representations cannot be directly linked to individual data points. This statistical anonymity aligns with the principles of differential privacy (DP), to create secure ML models but unlike DP, these do not provide quantifiable guarantees of privacy. \par

\section{Conclusion}
This study establishes the advantage of combining various learning strategies at the framework level to cater to robust real-world ML applications. Some of the key takeaways are: 
\begin{itemize}
    \item Model sharing can be seamlessly integrated into network exploration with minimal bandwidth overhead, making it an ideal first step for assessing feasibility among different strategies.
    \item Clients with limited data should solicit additional data from peers, leveraging VAEs for privacy-preserving synthetic data generation, which significantly enhances performance without overshadowing the original datasets.
    \item Employing synthetic data sharing as a supplementary measure post-convergence significantly boosts performance without compromising the integrity of the original datasets and works to revitalize model performance if it plateaus.
    \item In larger networks with high levels of non-I.I.D skews, learning from synthetic data is ideal for either balancing the data distribution or mitigating the bias inherent in noisy raw data.
\end{itemize} 
 Techniques like DP and secure multi-party computation (MPC) can complement the proposed data and model-sharing strategies for more robust privacy protection in sensitive settings. This work explored using partial and complete data or model exchange strategies in dynamic federated learning settings to combat the problem of knowledge transfer efficiency even in skewed data distributions. This underscores the need for learning strategies to adapt to their environments and participants in order to maximize effectiveness and ensure optimal outcomes.\par

\section*{Acknowledgments}

This research is supported by the National Research Foundation, Singapore under its Strategic Capability Research Centres Funding Initiative. Any opinions, findings and conclusions or recommendations expressed in this material are those of the author(s) and do not reflect the views of National Research Foundation, Singapore.

\bibliographystyle{named}
\bibliography{ijcai24}

\end{document}